\documentclass{article} \usepackage{aaai}
\usepackage{times}
\usepackage{epic}
\usepackage{ecltree}
\usepackage{graphicx}
\usepackage{latexsym}
\begin{document}

\newcommand{\set}{\mathcal}
\newcommand{\myset}[1]{\ensuremath{\mathcal #1}}

\renewcommand{\theenumii}{\alph{enumii}}
\renewcommand{\theenumiii}{\roman{enumiii}}
\newcommand{\figref}[1]{Figure \ref{#1}}
\newcommand{\tref}[1]{Table \ref{#1}}
\newcommand{\And}{\wedge}
\newcommand{\myldots}{.}

\newtheorem{mydefinition}{Definition}
\newtheorem{mytheorem}{Theorem}
\newtheorem{mytheorem1}{Theorem}
\newcommand{\myproof}{\noindent {\bf Proof:\ \ }}
\newcommand{\myqed}{\mbox{$\Box$}}

\newcommand{\mymod}{\mbox{\rm mod}}
\newcommand{\range}{\mbox{\sc Range}}
\newcommand{\roots}{\mbox{\sc Roots}}
\newcommand{\myiff}{\mbox{\rm iff}}
\newcommand{\alldifferent}{\mbox{\sc AllDifferent}}
\newcommand{\permutation}{\mbox{\sc Permutation}}
\newcommand{\disjoint}{\mbox{\sc Disjoint}}
\newcommand{\cardpath}{\mbox{\sc CardPath}}
\newcommand{\CARDPATH}{\mbox{\sc CardPath}}
\newcommand{\common}{\mbox{\sc Common}}
\newcommand{\uses}{\mbox{\sc Uses}}
\newcommand{\lex}{\mbox{\sc Lex}}
\newcommand{\usedby}{\mbox{\sc UsedBy}}
\newcommand{\nvalue}{\mbox{\sc NValue}}
\newcommand{\slide}{\mbox{\sc Slide}}
\newcommand{\SLIDE}{\mbox{\sc Slide}}
\newcommand{\circularslide}{\mbox{\sc Slide}_{\rm O}}
\newcommand{\among}{\mbox{\sc Among}}
\newcommand{\mysum}{\mbox{\sc MySum}}
\newcommand{\amongseq}{\mbox{\sc AmongSeq}}
\newcommand{\atmost}{\mbox{\sc AtMost}}
\newcommand{\atleast}{\mbox{\sc AtLeast}}
\newcommand{\element}{\mbox{\sc Element}}
\newcommand{\gcc}{\mbox{\sc Gcc}}
\newcommand{\gsc}{\mbox{\sc Gsc}}
\newcommand{\contiguity}{\mbox{\sc Contiguity}}
\newcommand{\PRECEDENCE}{\mbox{\sc Precedence}}
\newcommand{\precedence}{\mbox{\sc Precedence}}
\newcommand{\assignnvalues}{\mbox{\sc Assign\&NValues}}
\newcommand{\linksettobooleans}{\mbox{\sc LinkSet2Booleans}}
\newcommand{\domain}{\mbox{\sc Domain}}
\newcommand{\symalldiff}{\mbox{\sc SymAllDiff}}

\newcommand{\slidingsum}{\mbox{\sc SlidingSum}}
\newcommand{\MaxIndex}{\mbox{\sc MaxIndex}}
\newcommand{\REGULAR}{\mbox{\sc Regular}}
\newcommand{\regular}{\mbox{\sc Regular}}
\newcommand{\cfg}{\mbox{\sc Grammar}}
\newcommand{\CFG}{\mbox{\sc Grammar}}
\newcommand{\GRAMMAR}{\mbox{\sc Grammar}}

\newcommand{\DFA}{\mbox{\sc DFA}}
\newcommand{\STRETCH}{\mbox{\sc STretch}}
\newcommand{\SLIDEOR}{\mbox{\sc SlideOr}}
\newcommand{\NAE}{\mbox{\sc NotAllEqual}}

\newcommand{\todo}[1]{{\tt (... #1 ...)}}

\title{Decompositions of Grammar Constraints\thanks{The 
second author is funded by
the Australian Government's  Department of Broadband, Communications and the Digital Economy
and the 
Australian Research Council.}
}

\author{Claude-Guy Quimper\\
Ecole Polytechnique de Montr\'{e}al\\
Montr\'{e}al, Canada\\
claude-guy.quimper@polymtl.ca \And
Toby Walsh\\
NICTA and UNSW\\
Sydney, Australia\\
toby.walsh@nicta.com.au}
%\Date{}
%\setlength\titlebox{2in}

%\date{1st January 2008}

\maketitle
\begin{abstract}
A wide range of constraints can be 
compactly specified using automata or
formal languages. In a sequence of recent papers,
we have shown that
an effective means to reason with 
such specifications is to decompose
them into primitive constraints \cite{qwcp06,qwcp07}.
We can then, for instance, use 
state of the art SAT solvers
and profit from their advanced features
like fast unit propagation, clause learning, and
conflict-based search heuristics. 
This approach holds promise for solving 
combinatorial problems in scheduling, rostering,
and configuration, as well as problems in more diverse
areas like bioinformatics, software testing and natural language
processing. In addition, decomposition may be an effective
method to propagate other global constraints. 
\end{abstract}

\section{Introduction}

Constraint programming is 
an expressive and efficient technology
to solve a wide range of planning, scheduling, routing, and
configuration problems. 
However, constraint solvers are still some distance
from the ``model and run'' capability of solvers for mixed integer
programming (MIP) and propositional satisfiability (SAT). 
It requires considerable effort and expertise to
model a problem so that it can be solved using 
a constraint solver. A major
direction of research is therefore developing
new ways for the user to state their problem constraints that can then
be efficiently reasoned about automatically.

Tools from formal language theory (that is, 
automata and grammar rules) are useful
to specify many types of constraints. Recently, we 
have shown that decompositions
can be used to propagate such specifications
\cite{qwcp06,qwcp07}.
Our results demonstrate that such decompositions are
highly competitive with optimized code, 
but are easy to implement and benefit
from the current (and future) advances
in SAT solving technology. 
We believe that such methods may be of interest to researchers
in other areas like natural language processing, 
software testing, and bioinformatics where 
automata and grammar rules are already
used for problem specification
and where combinatorial problems naturally arise. 

\section{Background}
\newcommand{\sequenceLength}{\ensuremath{n}}

A constraint satisfaction problem (CSP) consists of a set of
variables, each with a finite domain of values, and a set of
constraints.
A constraint restricts values taken by some subset of
variables to a subset of the Cartesian product of their domains.  A
solution is an assignment of one value to each variable satisfying all
the constraints.  Systematic constraint solvers typically construct
partial assignments using backtracking search and 
constraint propagation 
to prune variable assignments which cannot be in any
solution. 

We will consider constraints which are specified
in terms of a grammar or automaton which accepts
just valid assignments for a sequence of variables. 
Regular languages are precisely those accepted by 
a deterministic finite automaton. 
A deterministic finite automaton (\DFA)
$\Omega$ is given by a 5-tuple $\langle Q,\Sigma,T,q_0,F\rangle$
where $Q$ is a finite set of states, $\Sigma$ is an alphabet,
$T:\Sigma \times Q \mapsto Q$ is the transition function, 
$q_0 \in Q$ is the initial state and $F \subseteq Q$ is the set of final
(or accepting) states. 
Context-free languages 
are above regular languages in the
Chomsky hierarchy. Context-free languages
are exactly those accepted by non-deterministic 
pushdown automaton
(that is, a 
automaton with a stack onto which we can push and
pop values, and a non-deterministic choice
of transitions). 
A context-free language
can be specified by a set of 
grammar rules in which the left-hand side has just
one non-terminal, and the right-hand side may have
a string of terminals and non-terminals. 
Any context-free grammar can be written
in Chomsky normal form in which each
rule yields either just one terminal or two non-terminals.

\section{An example}

Consider the shift-scheduling benchmark
introduced in~\cite{costregular}. 
The schedule of an employee in a company
is subject to the following rules. An employee either works on an
activity $a_i$, has a break ($b$), has lunch ($l$), or rests ($r$). When
working on an activity, the employee works for a
minimum of one hour. An employee can change activities after a break or
a lunch. A break is fifteen minute long and a lunch is one hour
long. Lunches and breaks are scheduled between periods of work. 

Employees can be part-time or full-time. A
part-time employee works at least three hours but less than six hours
a day and has one break. A full-time employee works between six and
eight hours a day and have a break, a lunch, and a break in that
order. Employees rest at the beginning and the end of the day. At some
time of the day, the business is closed and employees must either rest,
break, or have lunch. The day is divided into 96 time slots of 15
minutes. 
During time slot $t$, at least $d(t, a_i)$ employees must be
assigned to activity $a_i$. 
%Our goal is to minimize first
%the number of employees and then the number of hours worked. 

This is a complex problem but we can model 
it using a simple formal language. 
For each employee,
we introduce a sequence of 96 variables
(one per time slot) 
whose values must spell out a
string defined by the following grammar $G$.
\begin{eqnarray*}
& S \rightarrow R P R  \mid R F R & \\
R \rightarrow r R \mid r & L \rightarrow l L \mid l & A_i \rightarrow a_i 
A_i \mid a_i \\
W \rightarrow A_i & P \rightarrow W b W & F \rightarrow P L P
\end{eqnarray*}

\newcommand{\mytrue}{\emph{true}}
\newcommand{\myfalse}{\emph{false}}
\newcommand{\dom}{\ensuremath{\mbox{dom}}}

$S$ is the unique starting symbol. $R$ represents a period of rest.
$P$ represents a period of work by a part-time employee.
$F$ represents a period of work by a full-time employee.
$W$ represents a period of work on one activity. 
$L$ represents a lunch break. 

We also 
add restrictions on some of the productions. 
We attach
a Boolean function $f_N(i,j)$ to any non-terminal $N$
in a production where
$i$ represents the time period of the
start of the non-terminal and $j$ represents the 
length of the non-terminal. 
For example, with $W \rightarrow A_i$,
we have $f_W(i, j) \equiv j \geq 4$ since an employee works on an
activity for at least one continuous hour. In $F \rightarrow P L P$,
we have $f_L(i, j) \equiv (j = 4)$ since a lunch is one hour long. In
$S \rightarrow R P R$, we have $f_P(i, j) \equiv 13 \leq j \leq 24$
since a part-time employee works at least three hours and at most six
hours plus a fifteen minute break. In $S \rightarrow R F R$, we have
$f_F(i, j) \equiv 30 \leq j \leq 38$ which represents between six and
eight hours of work plus an hour and a half of idle time for the lunch
and the breaks. Finally, the productions $A_k \rightarrow a_k A_k \mid
a_k$ are constrained with $f_{A_k}(i, j) \equiv open(i)$ where
$open(t)$ returns \mytrue\ if $t$ is within business hours.
Such restrictions can greatly reduce the size of
the grammar needed but do not increase the asymptotic complexity
of reasoning about the \CFG\ constraint. 

When solving the problem with $m$ employees, the model consists of $m$
sequences subject to a constraint that
each spells out a string in this language. 
To ensure sufficient workers available for activity $a_i$
at time $t$, we also post the constraint that $\sum_j x(j, t, a_i)
\geq d(t, a_i)$ where $x(j, t, c)$ is an 0/1
variable set to 1 iff the $t^{th}$ character of
the $j$th sequence is $c$. 
As we argue in the next sections, decomposition into SAT
is an effective method to reason about constraints
specified in this way. 

\section{\regular\ constraint}

We start with one of the simplest but nevertheless
most useful methods so far proposed
for specifying constraints by means of a formal
language. The global constraint $\REGULAR([X_1,\ldots,X_n],\Omega)$
ensures that the values taken 
by a sequence of $n$ variables form 
a string accepted by the finite automaton $\Omega$ \cite{pesant1}. 
For example, consider the constraint that an employee starts
off with a period of rest ($r$), but once working ($w$) 
remains so until they again rest. 
We can specify this with the automaton
which starts in state $q_r$, 
stays in this state with $r$
but moves to state $q_{w}$ with $w$.
From $q_w$, the automaton stays
in this state with $w$ but moves
to state $q_{r'}$ with $r$. Once in 
state $q_{r'}$, only $r$ is accepted.
This defines the regular language $r^* w^* r^*$ 
which models this constraint on work and rests.

The \REGULAR\ constraint can be used to encode a wide variety
of useful global constraints like the \STRETCH\ constraint
\cite{pesant1} (which can specify constraints
on the length an employee works a continuous stretch
of shifts), and the 
\precedence\ constraint (which breaks value
symmetry) \cite{llcp2004,wecai2006}. 
The \REGULAR\ constraint can be decomposed
into a simple sequence of ternary constraints
\cite{qwcp06}. We merely need to
introduce finite domain variables, $Q_i$ which stand
for the state of the automaton after $i$ symbols,
and post ternary constraints
between variables representing neighbouring states 
to ensure appropriate transitions occur.
This decomposition is a highly efficient and effective
means to propagate the \REGULAR\ constraint. 
The decomposition 
does not hinder pruning which takes
$O(nd|Q|)$ time where $n$ is the length of the
sequence, $d$ is the domain size and $|Q|$ is
the number of states of the automaton.
This is asymptotically identical to the time complexity of 
the more complex monolithic propagator proposed
by Pesant based on dynamic programming \cite{pesant1}. 
More recently, Bacchus 
has proposed a SAT decomposition 
of the \REGULAR\ constraint \cite{bcp07}.
This can be seen as the SAT encoding of the
ternary transition constraints in our decomposition. 

Another advantage of our decomposition is that
we have explicit access to the states of the
automaton. Consider, for example, a
rostering problem where workers are allowed to work for 
up to three consecutive shifts and then must take a break. 
This can be specified with a simple \REGULAR\ language
constraint. Suppose 
we want to minimize the number of times a worker has to work for three
consecutive shifts. To model this, we can impose a 
global cardinality constraint on
the state variables to count the number of times
we visit the state representing
three consecutive shifts, and minimize
the value taken by this variable. 
It is much more complex to specify such 
an optimization constraint when the
states of the automaton are not represented
{\em explicitly} in the model. 

\section{Extensions of the \regular\ constraint}

Whilst deterministic finite automaton 
can in theory specify any type of constraint,
such specifications may not be compact. 
We therefore proposed a number 
of extensions including regular languages specified
by non-deterministic finite automata,
and soft and cyclic versions of the \REGULAR\ constraint
\cite{qwcp06}.
For instance, if a problem is over-constrained, we might
want to insist that we are ``near'' to a string
in the regular language. 
van Hoeve, Pesant and Rousseau have
proposed a generalization of the \REGULAR\ constraint to 
deal with such situations \cite{vanHoeve1}. 
\newcommand{\SoftREGULAR}{\mbox{$\mbox{\sc Regular}_{\rm soft}$}}
$\SoftREGULAR([X_1,\ldots,X_n],N,\Omega)$ 
holds iff the values taken by $X_1$ to $X_n$ form
a string that is at most distance $N$ from 
a string accepted by the \DFA\ given by $\Omega$. Distance is 
either Hamming distance (giving the usual 
variable-based costs) or edit distance (which
may be more useful in certain circumstances). 
In \cite{qwcp06}, we give encodings of
such soft \REGULAR\ constraints. 

As a second example,
we may want to find a repeating
sequence. We therefore introduced cyclic
forms of the \REGULAR\ constraint \cite{qwcp06}. 
In a rostering problem where the shift pattern
is repeated every four weeks, such a constraint can be used
to ensure that shifts changes only according to a set of
valid patterns (e.g. a night shift is only followed by another night
shift or a rest day, and is not followed by a day shift, even
at the end of the fourth week when we repeat back to the
first shift). 

\section{\cfg\ constraint}

Moving above regular languages in the
Chomsky hierarchy are context free
grammars. The \GRAMMAR\ constraint
\cite{grammar2,qwcp06} permits us to specify constraints using any
context-free grammar. 
Although context-free grammars are more complex
to reason about than regular
languages (e.g. parsing goes from $O(n)$ time for regular
languages to $O(n^3)$ time for context-free languages), 
they may compensate 
by requiring an exponentially smaller specification.
Since parsing (and propagation) depends linearly
on the size of the grammar, such reductions
in the size of the grammar can be of 
considerable benefit. 

Context-free grammar
constraints may have applications in a number of 
areas including:
\begin{description}
\item[Rostering and car sequencing:] to express constraints that
are not compactly expressible using a regular language as in
our earlier example;
\item[Configuration:]
to capture the hierarchically structure of 
a product (e.g. 
the computer consists of a motherboard, 
and input and output devices, 
the motherboard itself consists of a CPU
and memory, the CPU is an Intel or an AMD processor, 
etc.);
\item[Software verification:] to represent constraints
on the possible inputs to a 
  program for fuzz testing;
% Toby, shouldn't we cite the technical report of the people at Microsoft?
\item[Bioinformatics:]
to express patterns in genes and other types of sequences 
(e.g. context-free grammars are needed to represent
palindromes);
\item[Natural language processing:] 
to choose between different possible 
parsings.
\end{description}

In \cite{qwcp07,mipgrammar}, \CFG\ constraints have
been used to model complex shift-scheduling problems.
To reason about such \CFG\ constraints, we developed 
two propagators based on the CYK and Earley chart parsers
\cite{qwcp06}. Both use dynamic programming. 
Whilst the CYK propagator
takes $\Theta(n^3)$ time, the propagator based on the Earley chart
parser is just $O(n^3)$ and 
is not restricted to grammars in Chomsky 
normal form. 
%For instance, the Earley chart
%parser takes just $O(n^2)$ time on any unambiguous context-free
%grammar, and is linear on a wide range of useful grammars like LR(0).
More promising still, we have proposed 
a simple AND/OR decomposition based on the CYK parser 
which can be encoded into SAT \cite{qwcp07}.  We
have shown that this decomposition does not hinder propagation
and is asymptotically as fast as the
monolithic propagator based on the Earley chart
parser. To be more
precise, unit propagation on this decomposition will prune all
possible values in the same asymptotic time. 
Simpler grammars can also give a smaller decomposition. 
For instance, the decomposition is
just linear on a regular grammar. 

Decomposing global constraints in this way brings
several other advantages.  
First, we can easily add this global constraint
to any constraint solver. For example, we used the decomposition
to add the \GRAMMAR\ constraint to both a standard constraint toolkit
and a state of the art SAT solver. Second, decomposition gives an
efficient incremental propagator.  The solver can simply
wake up just those constraints containing variables whose
domains have changed, ignoring those parts of the decomposition 
that do not need to be propagated. Here, for example, we get
the first incremental
propagator for the \GRAMMAR\ constraint,
with a worst case cost down a whole branch 
of the search tree that is just the same as calling the
propagator once. 
Third, decomposition gives a propagator which we can backtrack
over efficiently. Modern SAT and CSP solvers use watch literals
so that we can backtrack one level up the search
tree in constant time. Fourth,
decomposition opens up a number of other possibilities which we are
only starting to explore. For example, it may make it easier to
construct no-goods, as well as cost measures for over-constrained
problems. Finally, a decomposition may make it easier to construct
constraint based branching heuristics.

\section{MIP encodings}

More recently, C\^{o}t\'{e}, Gendron, Quimper and Rousseau
have proposed mixed-integer programming (MIP) encodings
of the \REGULAR\ and \CFG\ constraints \cite{mipgrammar}. 
The MIP encoding of the \REGULAR\ constraint
introduces linear inequalities to model the
flow constructed by unfolding the automaton into
a layered transition graph. When this is the only
constraint in a problem, this can be solved with
a specialized path finding algorithm. However, when
there are other constraints in the problem, it can
be solved with a more general 0/1 MIP solver. 
The MIP encoding of the \CFG\ constraint introduces
linear inequalities which are derived from our
AND/OR decomposition of the CYK propagator. 
The MIP encoding has one significant difference.
If there is more than one parsing for a sequence,
it picks one arbitrarily whilst the CYK propagator
keeps all. This simplifies the MIP encoding without
changing the set of solutions since only one parsing is needed
to show membership in a context-free grammar. 
Experiments on a shift scheduling problem 
show that such MIP encodings are highly
competitive with other MIP formulations of
the problem. They open the door to specifying
complex shift scheduling rules using simple
tools from formal language theory, and solving
these problem with fast MIP solvers.

\section{Related work}

Vempaty introduced the idea of representing the solutions
of a CSP by a deterministic finite automaton \cite{vempaty1}. 
Such automaton can be used to answer questions
about satisfiability, validity and equivalence. 
Amilhastre generalized these ideas to non-deterministic automata,
and proposed heuristics to minimize the size of the automata
\cite{amilhastre1}. This approach was
then applied to configuration problems \cite{amilhastre2}. 
Boigelot and Wolper developed decision procedures for arithmetic
constraints based on automata \cite{boigelot1}.

Carlsson and Beldiceanu derived a propagation algorithm
for a chain of lexicographical ordering constraints
based on a deterministic finite automaton \cite{lexchain}. 
For the \REGULAR\ constraint, a propagation algorithm 
based on dynamic programming was given in \cite{pesant1}. Coincidently
Beldiceanu, Carlsson and Petit 
proposed specifying global constraints by
means of deterministic finite automaton
augmented with counters \cite{beldiceanu6}. 
Propagators for such automaton are constructed
automatically from the specification of the automaton
by constructing a conjunction of signature
and transition constraints. 
At the same time as \cite{qwcp06},
Sellmann 
proposed the \GRAMMAR\ constraint and gave a
monolithic propagator based on
the CYK parser \cite{grammar2}. 
Quimper and Rousseau \cite{quimperrousseau} used automta and context-free grammars as an operator for a large neighbourhood local search.
Finally, Golden and Pang propose the use of string variables 
which are specificed using regular expressions
or %finite 
automata and show how to enforce GAC on
matching, containment, cardinality and other constraints \cite{golden1}. 
\section{Conclusions}

Grammar constraints
specify that a sequence of 
variables are restricted to values 
spelling out a string within a given language. 
Such constraints are useful in a wide range of
scheduling, rostering and sequencing problems. 
We have shown that decomposition
is an efficient and effective method
to reason about such constraints
\cite{qwcp06,qwcp07}.
This an easy means to incorporate 
such grammar constraints into constraint toolkits
and SAT solvers. We believe that 
this approach holds promise 
for a wide range of other 
areas like bioinformatics and natural language
processing. 
Another promising direction is to learn 
grammar constraints from examples. 
We can, for instance, leverage on results and algorithms
from grammar induction. For example,
due to Gold's theorem,
it will not be possible 
to learn a \REGULAR\ constraint from just positive 
examples. Finally, decomposition into SAT may
prove effective for propagating other global
constraints.

\bibliographystyle{aaai}
%%\bibliographystyle{alpha}
%%\bibliography{/home/s5/tw/biblio/a-z,/home/s5/tw/biblio/pub}
%\bibliography{/home/tw/biblio/a-z,/home/tw/biblio/pub}
%%\bibliography{/n/endjinn/u6/tw/biblio/a-z,/n/endjinn/u6/tw/biblio/pub}
%%\bibliography{/usr/tw/biblio/a-z,/usr/tw/biblio/pub}
\bibliography{references.bib}

\begin{thebibliography}{} \itemsep=-0.5pt

\bibitem[\protect\citeauthoryear{Amilhastre, Fargier, \&
  Marquis}{2002}]{amilhastre2}
Amilhastre, J.; Fargier, H.; and Marquis, P.
\newblock 2002.
\newblock Consistency restoration and explanations in dynamic {CSPs} -
  application to configuration.
\newblock {\em Artificial Intelligence} 135:199--234.

\bibitem[\protect\citeauthoryear{Amilhastre}{1999}]{amilhastre1}
Amilhastre, J.
\newblock 1999.
\newblock {\em Representation par automate d'ensemble de solutions de
  probl\`{e}mes de satsifaction de contraintes}.
\newblock Ph.D. Dissertation, Universite Montpellier II / CNRS, LIRMM.

\bibitem[\protect\citeauthoryear{Bacchus}{2007}]{bcp07}
Bacchus, F.
\newblock 2007.
\newblock {GAC} via unit propagation.
\newblock In {\em Proc. of 13th Int. Conf. on Principles
  and Practice of Constraint Programming (CP2007)}.
%\newblock Springer.

\bibitem[\protect\citeauthoryear{Beldiceanu, Carlsson, \&
  Petit}{2004}]{beldiceanu6}
Beldiceanu, N.; Carlsson, M.; and Petit, T.
\newblock 2004.
\newblock Deriving filtering algorithms from constraint checkers.
\newblock In Wallace, M., ed., {\em Proc. of 10th Int.
  Conf. on Principles and Practice of Constraint Programming (CP2004)},
  107--122.
%\newblock Springer.

\bibitem[\protect\citeauthoryear{Boigelot \& Wolper}{2002}]{boigelot1}
Boigelot, B., and Wolper, P.
\newblock 2002.
\newblock Representing arithmetic constraints with finite automata: An
  overview.
\newblock In Stuckey, P., ed., {\em Proc. of the Int. Conf.
  on Logic Programming (ICLP 2002)},  1--19.
%\newblock Springer.

\bibitem[\protect\citeauthoryear{Carlsson \& Beldiceanu}{2002}]{lexchain}
Carlsson, M., and Beldiceanu, N.
\newblock 2002.
\newblock Arc-consistency for a chain of lexicographic ordering constraints.
\newblock Technical report {T2002-18}, Swedish Institute of Computer Science.

\bibitem[\protect\citeauthoryear{C\^{o}t\'{e} \bgroup \em et al.\egroup
  }{2007}]{mipgrammar}
C\^{o}t\'{e}, M.-C.; Gendron, B.; Quimper, C.-G.; and Roussean, L.-M.
\newblock 2007.
\newblock Formal languages for integer programming modeling of shift scheduling
  problems.
\newblock CIRRELT-2007-64, Centre interuniveritaire de recherche sur les r\'{e}seaux  d'enterprise, la logistique et la transport.

\bibitem[\protect\citeauthoryear{Demassey, Pesant, \&
  Rousseau}{2006}]{costregular}
Demassey, S.; Pesant, G.; and Rousseau, L.
\newblock 2006.
\newblock A cost-regular based hybrid column generation approach.
\newblock {\em Constraints} 11(4):315--333.

\bibitem[\protect\citeauthoryear{Golden \& Pang}{2003}]{golden1}
Golden, K., and Pang, W.
\newblock 2003.
\newblock Constraint reasoning over strings.
\newblock In Rossi, F., ed., {\em Proc. of 9th Int. Conf.
  on Principles and Practice of Constraint Programming (CP2003)},  377--391.
%\newblock Springer.

\bibitem[\protect\citeauthoryear{Law \& Lee}{2004}]{llcp2004}
Law, Y., and Lee, J.
\newblock 2004.
\newblock Global constraints for integer and set value precedence.
\newblock In {\em Proc. of 10th Int. Conf. on Principles
  and Practice of Constraint Programming (CP2004)},  362--376.
%\newblock Springer.

\bibitem[\protect\citeauthoryear{Pesant}{2004}]{pesant1}
Pesant, G.
\newblock 2004.
\newblock A regular language membership constraint for finite sequences of
  variables.
\newblock In Wallace, M., ed., {\em Proc. of 10th Int.
  Conf. on Principles and Practice of Constraint Programming (CP2004)},
  482--295.
%\newblock Springer.

\bibitem[\protect\citeauthoryear{Quimper \& Rousseau}{2007}]{quimperrousseau}
Quimper, C.-G., and Rousseau, L.-M.
\newblock 2007.
\newblock A large neighbourhood search approach to the multi-activity shift
  scheduling problem.
\newblock Technical Report CIRRELT-2007-56, 
Centre interuniveritaire de recherche sur les r\'{e}seaux  d'enterprise, la logistique et la transport.



\bibitem[\protect\citeauthoryear{Quimper \& Walsh}{2006}]{qwcp06}
Quimper, C.-G., and Walsh, T.
\newblock 2006.
\newblock Global grammar constraints.
\newblock In {\em 12th Int. Conf. on Principles and Practices of
  Constraint Programming (CP-2006)}.
%\newblock Springer. 
\newblock Longer version available as Technical Report COMIC-2006-005.

\bibitem[\protect\citeauthoryear{Quimper \& Walsh}{2007}]{qwcp07}
Quimper, C.-G., and Walsh, T.
\newblock 2007.
\newblock Decomposing global grammar constraints.
\newblock In {\em 13th Int. Conf. on Principles and Practices of
  Constraint Programming (CP-2007)}.
%\newblock Springer.

\bibitem[\protect\citeauthoryear{Sellmann}{2006}]{grammar2}
Sellmann, M.
\newblock 2006.
\newblock The theory of grammar constraints.
\newblock In {\em Proc. of 12th Int. Conf. on Principles
  and Practice of Constraint Programming (CP2006)},  530--544.
%\newblock Springer.

\bibitem[\protect\citeauthoryear{van Hoeve, Pesant, \&
  Rousseau}{2006}]{vanHoeve1}
van Hoeve, W.-J.; Pesant, G.; and Rousseau, L.-M.
\newblock 2006.
\newblock On global warming : Flow-based soft global constaints.
\newblock {\em Journal of Heuristics}.
\newblock 12(4-5): 347-373.

\bibitem[\protect\citeauthoryear{Vempaty}{1992}]{vempaty1}
Vempaty, N.~R.
\newblock 1992.
\newblock Solving constraint satisfaction problems using finite state automata.
\newblock In {\em Proc. of the 10th National Conf. on AI},
  453--458.
\newblock AAAI.

\bibitem[\protect\citeauthoryear{Walsh}{2006}]{wecai2006}
Walsh, T.
\newblock 2006.
\newblock Symmetry breaking using value precedence.
\newblock In {\em Proc. of the 17th ECAI}.
\newblock European Conf. on AI.

\end{thebibliography}
%%\bibliography{/home/arp/disk1/tw/biblio/a-z,/home/arp/disk1/tw/biblio/pub}
%%\bibliography{/u6/tw/biblio/a-z,/u6/tw/biblio/pub}
%%\bibliography{/usr/local/users/tw/biblio/a-z,/usr/local/users/tw/biblio/pub}
%\bibliography{biblio}

%\end{thebibliography}

\end{document}